\documentclass[letterpaper, 10 pt, conference]{ieeeconf}
\IEEEoverridecommandlockouts

\usepackage{graphicx}
\usepackage{cite}
\usepackage{amsmath}
\usepackage{amssymb}
\usepackage{algorithm}
\usepackage{hyperref}
\usepackage{algpseudocode}
\usepackage{booktabs}
\usepackage{xcolor}

\title{\LARGE \bf
LERa: Replanning with Visual Feedback in Instruction Following
}

\author{Svyatoslav~Pchelintsev$^{1*}$, Maxim~Patratskiy$^{1*}$, Anatoly~Onishchenko$^{1}$, Alexandr~Korchemnyi$^{1}$, \\Aleksandr~Medvedev$^{2}$, Uliana~Vinogradova$^{2}$, Ilya~Galuzinsky$^{2}$, Aleksey~Postnikov$^{2}$, \\Alexey~K.~Kovalev$^{3,1}$, and Aleksandr~I.~Panov$^{3,1}$
\thanks{*Svyatoslav~Pchelintsev and Maxim~Patratskiy contributed equally to this work. $^{1}$MIPT, Dolgoprudny, 141701, Russia $^{2}$Sberbank of Russia, Robotics Center, Moscow, 117997, Russia $^{3}$AIRI, Moscow, 121170, Russia {\tt\small \{kovalev,panov\}@airi.net}}
}

\begin{document}

\maketitle
\thispagestyle{empty}
\pagestyle{empty}

\begin{abstract}
Large Language Models are increasingly used in robotics for task planning, but their reliance on textual inputs limits their adaptability to real-world changes and failures. To address these challenges, we propose LERa~---~\underline{L}ook, \underline{E}xplain, \underline{R}epl\underline{a}n~---~a Visual Language Model-based replanning approach that utilizes visual feedback. Unlike existing methods, LERa requires only a raw RGB image, a natural language instruction, an initial task plan, and failure detection~---~without additional information such as object detection or predefined conditions that may be unavailable in a given scenario. The replanning process consists of three steps: (i) Look~---~where LERa generates a scene description and identifies errors; (ii) Explain~---~where it provides corrective guidance; and (iii) Replan~---~where it modifies the plan accordingly. LERa is adaptable to various agent architectures and can handle errors from both dynamic scene changes and task execution failures. We evaluate LERa on the newly introduced ALFRED-ChaOS and VirtualHome-ChaOS datasets, achieving a 40\% improvement over baselines in dynamic environments. In tabletop manipulation tasks with a predefined probability of task failure within the PyBullet simulator, LERa improves success rates by up to 67\%. Further experiments, including real-world trials with a tabletop manipulator robot, confirm LERa’s effectiveness in replanning. We demonstrate that LERa is a robust and adaptable solution for error-aware task execution in robotics. 
The project page is available at \url{https://lera-robo.github.io.}
\end{abstract}

\section{INTRODUCTION}
\label{sec:intro}
Large Language Models (LLMs) trained on Internet-scale data can solve problems that they were not originally designed for~\cite{kojima2022large}. This has led to the widespread use of LLMs in robotics for task planning~\cite{huang2022language,kovalev2022application,sarkisyan2023evaluation,10160591,10801328,onishchenko2025lookplangraph}.

However, there is no guarantee that a robot will successfully execute a generated task plan in a real environment. This is because LLMs rely on common sense knowledge acquired during training and do not account for dynamic environmental changes or task failures. For example, given the instruction \textit{``Heat a slice of pizza,''} the model may well generate a plan that includes the task \textit{``Open the microwave.''} However, if the microwave is already open, perhaps because the previous user forgot to close it, the plan will result in an error, preventing successful execution. Similarly, if the pizza slice falls while being placed inside, the robot may continue executing the plan without recognizing the failure, leading to an incomplete or incorrect outcome.

To address this problem, various LLM-based replanning approaches have been proposed~\cite{song2023llm,guo2024doremi,joublin2024copal}. However, LLMs rely solely on textual input, which does not fully capture the state of the environment and requires this state to be represented in textual form, such as a list of objects or a scene description~\cite{huang2022language,10161317,10160591,10801328,song2023llm}. This textual representation is typically derived from the output of computer vision systems, such as object detectors or semantic segmenters. Visual Language Models (VLMs), on the other hand, can directly process both visual and textual modalities, making them more suitable for capturing the state of the environment and enabling replanning. However, most modern VLM-based replanning approaches: (i) are designed for a specific task planner or agent architecture, making them difficult to adapt to different architectures~\cite{zhang2023grounding,yang2024guidinglonghorizontaskmotion,mei2024replanvlm,skreta2024replan}; (ii) require not only information from the RGB camera and the instruction with the corresponding task plan for replanning but also additional data, such as pre- and post-action conditions, target object positions, and object bounding boxes~---~information that may not be available in a given scenario~\cite{yang2024guidinglonghorizontaskmotion,mei2024replanvlm}; (iii) are primarily designed to handle a single type of error, limiting their applicability to other error types~\cite{zhang2023grounding,yang2024guidinglonghorizontaskmotion,skreta2024replan}.
\begin{figure}
    \centering
 \includegraphics[width=1\columnwidth]{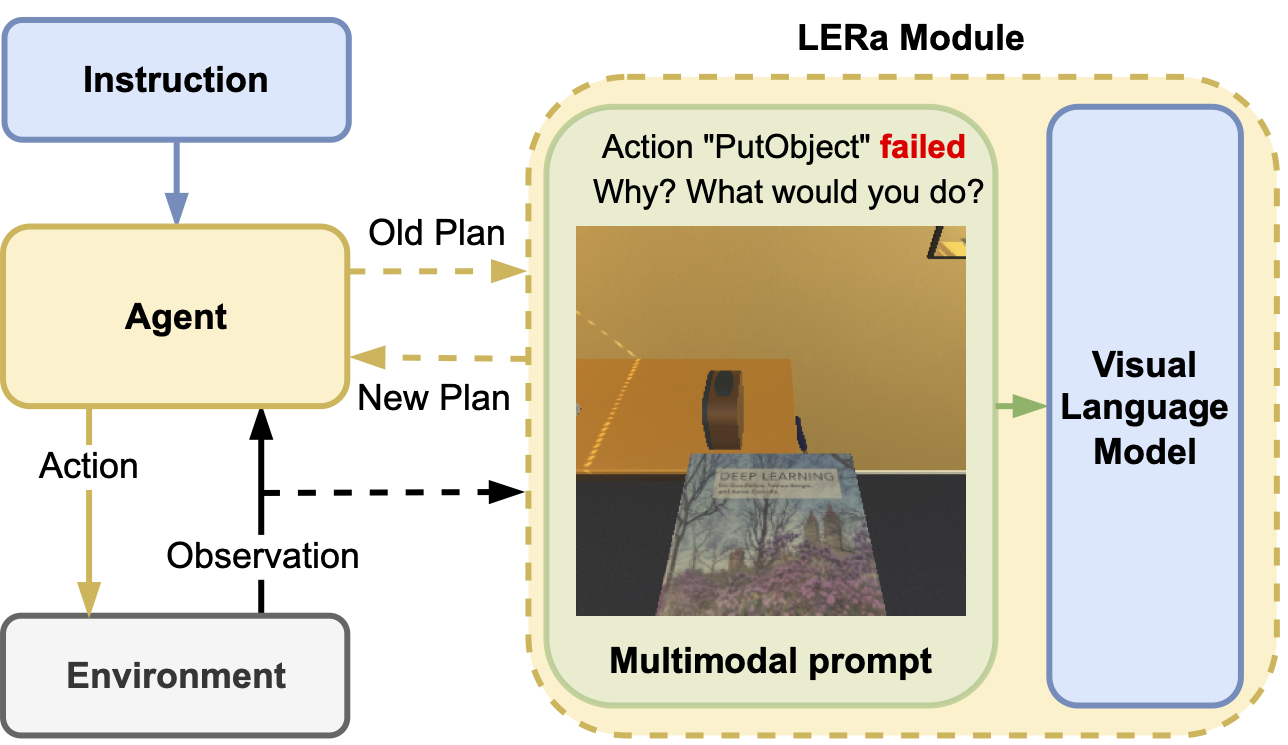}
    \caption{
    An approach without error handling can be enhanced by the LERa module to dynamically update the plan (a queue of tasks) when errors occur during plan execution. On the left (solid arrows), the plan is generated once at the beginning of an episode and remains unchanged throughout. With LERa (dashed arrows), a Visual Language Model is used to replan based on visual feedback whenever an error occurs.
    }
    \label{fig:visab}
    \vspace{-10pt}
\end{figure}

In this paper, we address these problems by proposing \textbf{LERa}~---~\textbf{\underline{L}}ook, \textbf{\underline{E}}xplain, \textbf{\underline{R}}epl\textbf{\underline{a}}n~---~a VLM-based approach for replanning with visual feedback. LERa uses only basic information for replanning: (i) a raw RGB camera image, without requiring detected objects, masks, or any other additional information; (ii) failure-triggering action, without requiring an explanation of its cause; (iii) the initial natural language instruction; and (iv) the task plan. LERa makes no specific assumptions about the agent’s architecture and can extend any agent that detects action execution failures and requests replanning (see Fig.~\ref{fig:visab}). Moreover, LERa is capable of handling errors caused by dynamic environmental changes and action execution failures.
The replanning process in LERa consists of three steps. \textbf{Look}~---~given the raw RGB camera image and information about the action failure, LERa generates a scene description and identifies the cause of the error. \textbf{Explain}~---~based on the identified information, LERa explains how to correct the plan and addresses the replanning problem without being constrained to a specific output format. \textbf{Replan}~---~LERa modifies the old plan by incorporating the proposed changes and passes the final version of the plan to the agent for execution.

To evaluate LERa's ability to replan, we collected the \textbf{ALFRED-ChaOS} and \textbf{VirtualHome-ChaOS} datasets, each containing more than 500 episodes, based on ALFRED~\cite{shridhar2020alfred} and VirtualHome~\cite{puig2018virtualhome} environments. 
LERa improves success rates by over 40\% compared to baselines, including a one-step VLM-based replanner. We also test LERa on 10 tabletop manipulation tasks in PyBullet~\cite{coumans2021} with stochastic action failures. LERa achieves up to 67\% higher success than a no-replanning baseline.

We also conducted extensive ablation experiments that confirmed the necessity of all three steps in the replanning process. Additionally, we performed experiments with different VLMs that demonstrated how VLM quality affects replanning performance. Further experiments with a non-ideal subtask checker on ALFRED-ChaOS showed the consistent superiority of LERa when working with imperfect error detection. To demonstrate the generalizability of our approach to real-world scenarios, we integrated LERa into the control system of a real robot for tabletop manipulation. LERa successfully handled replanning in 15 out of 18 trials.

Our main contributions can be summarized as follows:
\begin{enumerate}
    \item \textbf{We propose LERa}, a VLM-based replanning approach that utilizes visual feedback, makes minimal assumptions about the agent's architecture, and is capable of replanning for both dynamic scene changes and action execution errors.
    \item \textbf{We introduce three different environments} based on well-known open-source simulators, which can be used to evaluate replanning approaches.
    \item \textbf{We conduct rigorous experiments and ablations across various environments}, demonstrating that LERa successfully handles replanning under different conditions and for different types of errors.
    \item \textbf{We validate the applicability of LERa to real-world problems} by integrating it into a real robot control system for tabletop manipulation, showing that LERa effectively handles replanning.
\end{enumerate}

\section{RELATED WORK}
VLM-based approaches are widely used in robotics for replanning, but they usually either rely on additional information beyond the camera image, assume a specialized agent architecture, or replan only for a specific type of error. In~\cite{hu2023look}, the VLM is used as the initial task planner, and the replanning problem is not structurally separated. Although this approach significantly simplifies the architecture, it does not account for specific features of the environment or the robot in the initial planning~\cite{yang2024guidinglonghorizontaskmotion}.
In~\cite{skreta2024replan}, a complex interaction structure is used between planning modules at different levels, where VLM is employed to confirm the correctness of an action choice before its execution in the environment. In~\cite{zhang2023grounding}, replanning is framed as a visual question-answering problem, handling only errors related to action failures and requiring information about the preconditions and effects of actions.
In~\cite{mei2024replanvlm}, a VLM-based Extra Bot is used as an action failure detector by analyzing frames before and after an action is performed. Replanning is handled by the Design Bot, which also functions as a task planner, imposing constraints on the agent's architecture.
In~\cite{yang2024guidinglonghorizontaskmotion}, VLM is used to generate intermediate goals instead of actions during replanning. In this case, information about detected objects, their relationships, and collisions is provided to VLM along with the image. In~\cite{duan2024aha}, VLM is further trained to detect and explain errors.

Our approach differs from all these methods in that LERa uses only the minimum information necessary for replanning and does not make specific assumptions about the planner architecture. This improves its generalizability and simplifies its use. It can handle multiple types of errors and does not require further training of the VLM.

\section{METHOD}
\label{sec:method}
We propose LERa, a VLM-based approach for replanning (Section~\ref{subsec:problem}) with visual feedback. LERa makes no specific assumptions about the agent's architecture and can be used with any agent that provides action failure detection and replanning request capabilities (Section~\ref{subsec:lera}). It employs a three-step prompting strategy (Section~\ref{subsec:prompting}) that allows it to handle different types of errors without requiring additional information. LERa relies solely on an RGB image, action failure detection information, an initial natural language instruction, and a task plan.

\subsection{Problem Formulation}
\label{subsec:problem}
We consider the problem of \textit{replanning} as the problem of adjusting an initial plan $P$, which consists of a sequence of actions (tasks) $P=(a_1, a_2, \dots, a_n)$ (where $n$ is the plan length) executable in the environment, upon receiving information about the occurrence of an error during execution of an action (task). We do not consider low-level actions, which result from motion planning rather than task planning, to be components of the plan. Therefore, we use the terms \textit{``action''} and \textit{``task'}' synonymously where this does not cause confusion.
By \textit{plan adjustment}, we mean modifying the sequence of actions in the plan $P=(a_1, a_2, \dots, a_n)$ to $P'=(a'_1, a'_2, \dots, a'_{n'})$ so that execution can continue successfully in the environment. Since we do not address the problem of initial planning, we assume that the plan $P$ always leads to the achievement of the goal in the absence of environmental changes or unsuccessful action execution.

We also assume that a replanner $R$: 1) operates only on information available after the planning stage~---~namely, a natural language instruction $I$, the task plan $P$, and information about the occurrence of an error at time $t$ (denoted as $E_t$ and obtained using a task checker $TC$); 2) has access to the current raw RGB observation $O_t$.
Thus, the problem is reduced to creating a function $R$ that, given all the described inputs, generates the corrected plan $P'$: $R(O_t, E_t, I, P) = P'$.

\subsection{The LERa Approach}
\label{subsec:lera}
\begin{figure*}[t]
  \centering
  \includegraphics[width=1.0\linewidth]{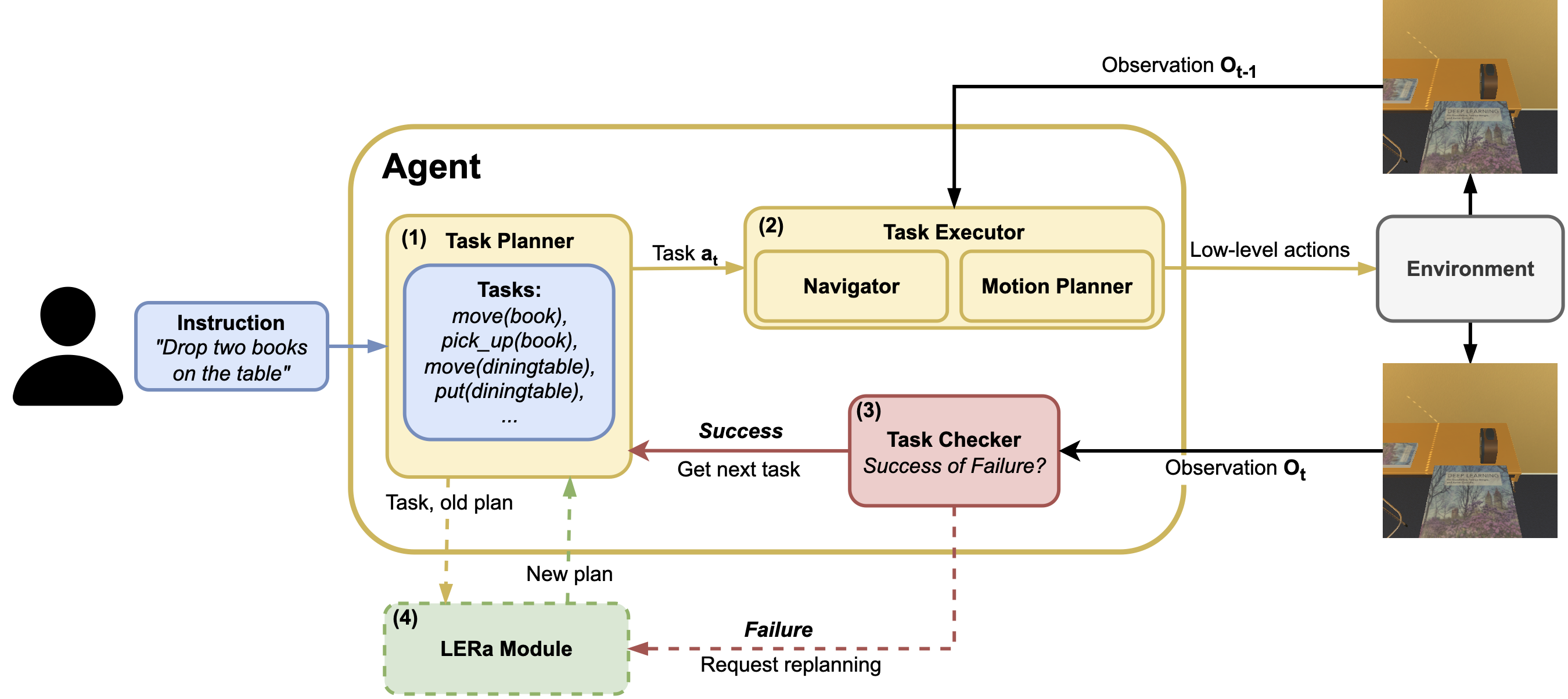}
  \caption{The supposed structure of an agent to use LERa consists of three modules: (1) Task Planner, (2) Task Executor, and (3) Task Checker. \textcolor{blue}{Blue} is for language, \textcolor{yellow}{yellow} is for planning and execution, \textcolor{red}{red} is for self-checking, and \textcolor{green}{green} is for replanning.}
  \label{fig:method_overview}
  \vspace{-10pt}
\end{figure*}

LERa makes fairly general assumptions about an agent's architecture, requiring three main modules: 1)~a Task Planner, 2)~a Task Executor, and 3)~a Task Checker (Fig.\ref{fig:method_overview}).
After the Task Planner (1) has generated a task plan $P$ from the language instruction $I$ and, possibly, the environment initial state, the workflow is as follows: (i)~the Task Planner (1) manages task execution by assigning the next action $a_i$; (ii)~the Task Executor (2) produces low-level actions, potentially incorporating navigation and motion planning; (iii)~the Task Checker (3) verifies task execution using environment feedback, then either requests the next action $a_{i+1}$ if execution is successful or triggers replanning by LERa (4) if an error occurs.
Due to its modular structure, LERa does not depend on how the initial plan is generated. It requires only that the input to the replanner be a sequence of actions to revise, enabling integration with various high-level planning systems.

\begin{figure}[t]
    \centering
\includegraphics[width=\linewidth]{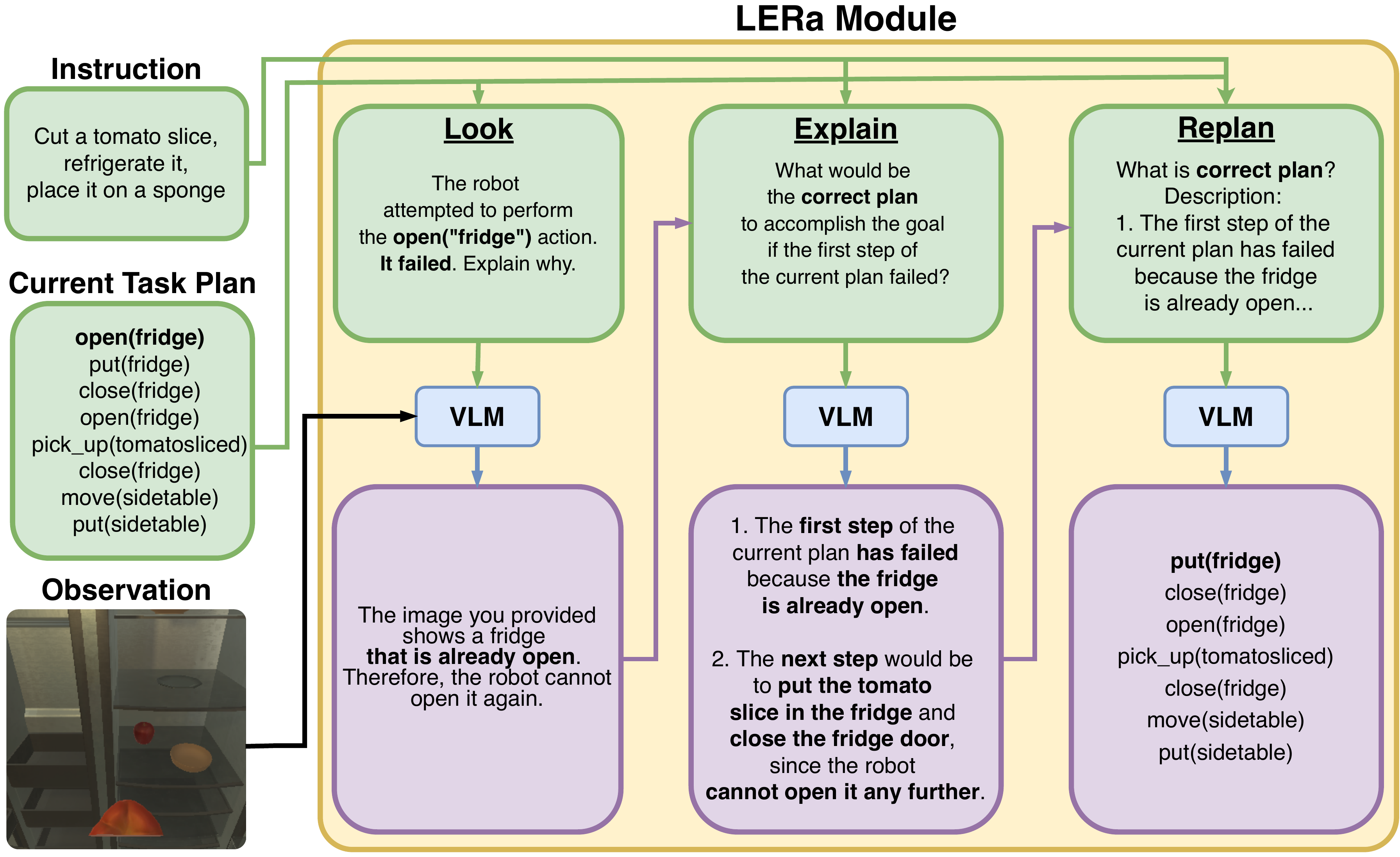}
\vspace{-10pt}
    \caption{A step-by-step example of replanning using the LERa module. It consists of three main steps: Look, Explain, and Replan. Each step solves its own problem in order to modify the current plan $P$ to a new plan $P'$.}
    \label{fig:replan}
    \vspace{-15pt}
\end{figure}

The LERa module updates the agent’s plan $P$ to a revised plan $P'$, mitigating execution errors.
It operates in three main steps: \textit{Look}, \textit{Explain}, and \textit{Replan}. The algorithm for LERa's operation is presented in Algorithm~\ref{alg:alg}, and Fig.~\ref{fig:replan} illustrates a step-by-step example of the replanning process.

\textbf{Look Step:} At the moment of the execution failure of a task $a_t$ ($TC(a_t) = 0$), the agent captures the visual observation $O_t$. Based on $O_t$ and $a_t$, LERa creates a scene description and provides various failure reasons. The predicted description is denoted by $L$.

\textbf{Explain Step:} Given the current plan $P$, the failed action $a_t$, and the information from the previous step $L$, LERa outputs a textual description of the error and a conceptual sequence of new actions -- $E$.

\textbf{Replan Step:} At this step, LERa constructs a new action plan $P' = (a_t', a_{t+1}', \dots, a_n')$. The new plan is generated without any additional explanation of its applicability or validation of execution feasibility.

\begin{algorithm}[h]
\caption{LERa Replanning Process}
\begin{algorithmic}[1]
\Require \\
    Natural language instruction $I$, \\
    Current observation $O_t$, \\
    Current task plan $P = (a_t, ...,a_n).$
\Ensure Updated action plan $P'$ that ensures successful task completion
\State $L \leftarrow \text{Look}(O_t, a_t)$ \Comment{Generate scene description and identify failure cause}
\State $E \leftarrow \text{Explain}(I, L, P)$ \Comment{Explain error and propose conceptual plan}
\State $P' \leftarrow \text{Replan}(I, E, P)$ \Comment{Predict updated sequence of actions}
\State \textbf{return} $P'$
\end{algorithmic}
\label{alg:alg}
\end{algorithm}
\vspace{-10pt}

\subsection{Prompting Strategies}
\label{subsec:prompting}
\begin{figure}[h]
    \centering
    \includegraphics[width=0.9\columnwidth]{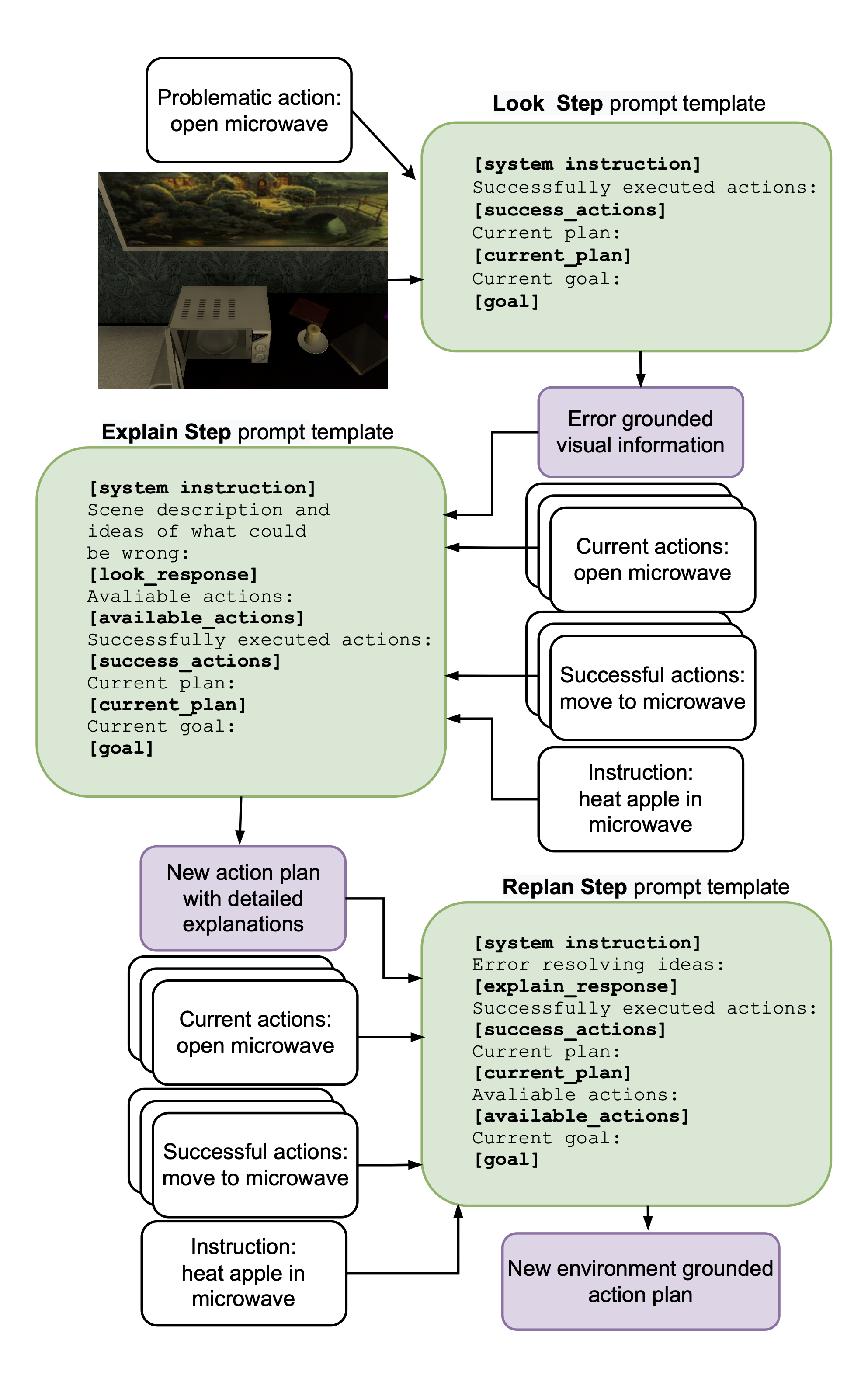}
    \caption{Prompt templates used by the LERa module.}
    \label{fig:prompt_templates}
\end{figure}

The LERa method utilizes three different prompt templates~---~one for each step~---~to solve specific tasks (Fig.~\ref{fig:prompt_templates}). In the \textbf{Look Step}, LERa identifies the causes of the error and provides simple ideas for a solution. To prevent the VLM from being overloaded with unnecessary information, only a limited set of key facts about the environment is included in the system instruction. For the same reason, only the first step of the current plan is provided to the model along with the observation. The \textbf{Explain Step} is designed to address the replanning task without being constrained by any output format. The objective of this stage is to reason about the given task instruction, the existing plan, and the newly acquired visual information in order to generate an updated task plan. In addition, the model is required to provide an explanation for each predicted step, ensuring that the output contains as much relevant information as possible to support the final replanning step. The \textbf{Replan Step} constructs the final corrected task plan. Its prompt includes few-shot examples to constrain formatting, a list of available environment actions with descriptions, the existing task plan, and the newly predicted task plan with step explanations. This step ensures that the new plan is executable in the virtual environment, as the Explain Step may produce infeasible actions. Additionally, the Replan Step refines the plan to align it with the environment or to correct logical errors.

\section{EXPERIMENTS}
\label{sec:experiments}
\begin{figure}[ht]
    \centering
    \includegraphics[width=\linewidth]{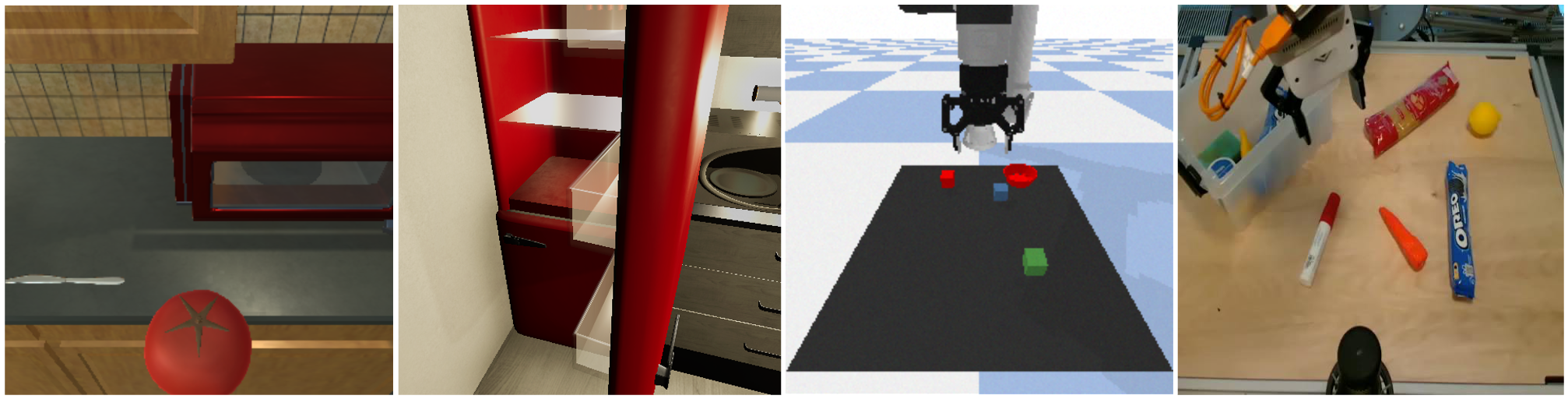}
    \caption{We evaluate LERa across diverse set of environments. From left to right, visual observations from: ALFRED, VirtualHome, TableTop PyBullet, and TableTop Robotic Stand.
    \vspace{-30pt}
    }
    \label{fig:all-envs}
\end{figure}
To demonstrate the generalizability of the LERa approach, we evaluated it in diverse simulation environments and on a real tabletop manipulation robot (Fig.~\ref{fig:all-envs}). We conducted experiments with different parts of LERa to assess the contribution of each component to overall success. Additional evaluations focused on measuring the impact of the VLM on the overall success rate. Moreover, we tested LERa with a non-ideal checker to evaluate its ability to correctly assess the environment state and restore the original plan when no actual errors occurred. In addition, we conducted experiments on a real robot to demonstrate the applicability of the method in real world conditions.

\subsection{Environments}
We conducted a diverse set of experiments in three virtual environments: ALFRED~\cite{shridhar2020alfred}, VirtualHome~\cite{puig2018virtualhome} and PyBullet. In ALFRED and VirtualHome, the experiments focused on assessing the approach’s ability to handle errors caused by unexpected object state changes. In PyBullet, we evaluated its effectiveness in scenarios where the last executed action completed but the intended goal was not achieved. Such tasks require the agent to detect partial failure, infer its cause, and adjust the plan accordingly.

\textbf{ALFRED-ChaOS.}
In ALFRED, an embodied agent is asked to complete a household task, e.g. ``Put a clean sponge on a metal rack'', by following low-level instructions, e.g. ``Go to the left and face the faucet side of the bathtub. Pick up left most \dots''. We took episodes from ALFRED's ``valid\_seen'' and ``valid\_unseen'' splits, filtered out the repetitive ones as they corresponded to the same ground truth (GT) plans, and changed states of objects \textbf{that affect the execution of the task}. As a result, we obtained 251 episodes with 290 cases of objects with changed states. We combined these episodes with their original versions and ended up with a total of 502 episodes. We refer to the resulting dataset as \textbf{ALFRED-ChaOS}, which stands for \textbf{Cha}nged \textbf{O}bject \textbf{S}tates.

\textbf{VirtualHome-ChaOS.}
VirtualHome is similar to ALFRED as an embodied agent is asked to complete household tasks.
In this environment, we adapted three types of tasks: heating objects in a microwave, placing objects in a refrigerator, and washing objects in a dishwasher. Two distinct tasks corresponding to each task type were collected for each of the 50 scenes, ensuring variation in the involved objects. Then, scenes that lacked essential containers for the tasks were excluded from the set.
Using this approach, we created 262 tasks requiring interactions with various objects in the scene. To implement changing in objects states, we modified the states of key containers relevant to each task. Combining these modified tasks with the original ones resulted in a total of 524 tasks. We refer to the resulting dataset as \textbf{VirtualHome-ChaOS}.

\textbf{TableTop PyBullet.}
In ALFRED and VirtualHome, we evaluated replanning capabilities in scenarios where the objects' states differed from the initial states. To evaluate the ability to resolve errors that occur during action execution, we designed a series of experiments in PyBullet. The agent has two primary actions: ``pick'' and ``place'', but may drop the object during their execution with a predefined probability. We prepared 10 tasks and run each 10 times with different object's locations and drop moments, resulting in 100 scenarios. Each task has a different plan length and difficulty. For example, ``Place red block in red bowl'' and ``Build two towers in bowls: blue on red, yellow on green'' require 4 and 16 actions, respectively. In the first task, the agent must perform the actions ``locate red block'', ``pick red block'', ``locate red bowl'', and ``place red bowl'', whereas the plan in the second task is four times longer, as the agent has to move four blocks instead of one.

\textbf{TableTop Robotic Stand.}
To evaluate LERa in a real-world environment, we used the XArm6 robotic platform, which was equipped with XArm grippers and RealSense L515 and D435 sensors. It performed manipulation tasks with two primary actions: ``pick'' and ``place''. The planner, based on an open-source LLM, generated a plan using object recognition data, while the fine-tuned VLM-based checker evaluated the success of each action execution. Replanning was triggered when the checker detected execution errors, such as misplacement or misclassification of objects.
The experiments consisted of 18 different tasks involving the manipulation of various sets of objects, moving them from a table to a container. Notably, in this real-world setting, errors frequently occurred due to inaccuracies in the gripper position predicted by the agent. This led to failures in grasping or placing objects. Additionally, some episodes involved initial misclassification by the visual detector, resulting in an incorrect initial plan and, consequently, the inability to correctly execute the human instruction.

\subsection{Metrics}
We use the following metrics to evaluate replanning quality: \textbf{Success Rate} (\textbf{SR}), \textbf{Goal-Condition Success} (\textbf{GCR}), and \textbf{Success Rate of Replanning} (\textbf{SRep}).  \textbf{SR} is the ratio of the number of successfully completed episodes to the total number of episodes. An episode is considered successfully completed when all target objects are in their target positions and states at the end of the episode. For ALFRED and VirtualHome,  this means all objects are in the correct places and in the specific states, e.g., the apple is cut and it is in the closed fridge. For PyBullet, the target states are the correct positions of blocks in bowls, e.g., the blue block is on the red one and both are in the yellow bowl. \textbf{GCS} is the ratio of the number of goal conditions successfully met at the end of an episode to all goal conditions in that episode, averaged over the number of episodes. This metric can be interpreted as the fraction of all objects that met their target state. \textbf{SRep} is the ratio of the number of successful replanning attempts to the total number of replanning attempts in an episode, averaged over the number of episodes. Replanning is considered successful if it results in the agent continuing to execute the plan and resolving the issue.

\subsection{Agents}
The LERa approach functions solely as a replanning module, meaning it cannot be used for initial plan generation with guarantees of effectiveness. To evaluate the replanning capability of LERa and isolate its replanning performance from errors in other modules, we used ground-truth plans and oracle agents.
Oracle agents are those that have perfect or near-perfect knowledge of the environment and execute actions with the maximum achievable probability of success (usually 100\%).

We used oracle agents for each virtual environment in our experiments and the agents that were not able to replan are denoted as ``Oracle''. These agents were provided with a complete, predefined initial plan that solves the task assuming there are no objects with altered states or execution errors.
Therefore, if an action in the plan cannot be executed, the Oracle agents simply moved on to the next action, eventually leading to a failed episode.
In the VirtualHome and PyBullet environments, the agents achieved a 100\% success rate (SR) as long as there were no changes in the scenes.
In the ALFRED environment, the Oracle agent successfully completed tasks with an SR of approximately 75\%. Although the GT plans were used, ALFRED’s predefined templates are suboptimal, as they do not account for objects hidden in receptacles. For example, if the task is to retrieve a knife from one of multiple drawers, Oracle may teleport to an empty one. These imperfections reduced SR by approximately 15–20\%. An additional $\approx 5\%$ drop resulted from AI2THOR bugs, such as unexplained \textit{``PutObject''} failures or objects being rendered without corresponding masks.

\subsection{Baselines}
To assess the contribution of each step in LERa, we conducted experiments with its different variations -- \textbf{Ra}, \textbf{LRa}, and \textbf{ERa} -- each representing a distinct combination of replanning steps. These experiments allow for a detailed analysis of the role of each component and provide a structured comparison against the full method. Each LERa variation is an exact replica of the corresponding part of the main approach, ensuring consistency in evaluation. The ERa variant, in particular, represents one of the most commonly used error correction strategies. It does not process visual information but instead relies on textual feedback from the environment that explicitly indicates whether a given action was unsuccessful. This makes ERa a widely applicable baseline as it simulates a scenario in which the agent corrects errors based solely on system-provided failure signals rather than on direct perception.

\textbf{Baseline} represents a method commonly used in recent research~\cite{zhang2023grounding,mei2024replanvlm}. It is not a direct implementation of any specific approach, but rather an abstraction of the one-shot plan prediction paradigm in which the VLM receives all available information in a single step.

The method immediately replans upon receiving visual information and additional task-related data, directly outputting a corrected plan. In our implementation, it corresponds to the ``Replan'' step of LERa, with the addition of visual feedback. However, unlike LRa, Baseline operates in a single-step manner, generating a new plan without intermediate reasoning.

\subsection{Ablations}
In the additional experiments, we demonstrate how the LERa module performs when other modules make mistakes. In real-world scenarios, it is nearly impossible to create a checker that detects errors with absolute accuracy. Thus, the first three agents (``O-05'', ``O-10'' and ``O-15'') have the imperfect Task Checker that flips the true prediction with probabilities of 0.05, 0.10 and 0.15, respectively. The fourth agent (``O-FC'') uses FILM's \cite{min2021film} Task Checker that has a check error probability of $\approx 2.5\%$.

\subsection{Implementation Details}
We used public APIs for experiments with proprietary models (gpt-4o and gpt4o-mini~\cite{hurst2024gpt}, Gemini-Flash-1.5 and Gemini-Pro-1.5~\cite{team2023gemini}) and two NVIDIA RTX Titan 26 GB for experiments with open-source models (MiniGPT-v2~\cite{chen2023minigptv2}, LLaMA-3.2~\cite{dubey2024llama}): one GPU handled a VLM, and the other managed a virtual environment and an agent.
MiniGPT-v2 was used in experiments with imperfect checkers.
The generation hyperparameters were configured to ensure deterministic responses.
VirtualHome and PyBullet do not require significant computational resources and were therefore run on a local PC.

\section{RESULTS}

\begin{table*}[t]
\caption{
Replanning results for different agents and environments.}
\centering
\small
\resizebox{0.9\textwidth}{!}{
\begin{tabular}{lccc|ccc|ccc|ccc|ccc}
    \toprule
    & \multicolumn{3}{c}{\textbf{ALFRED-ChaOS (Seen)}} 
    & \multicolumn{3}{c}{\textbf{ALFRED-ChaOS (Unseen)}} 
    & \multicolumn{3}{c}{\textbf{VirtualHome-ChaOS}} 
    & \multicolumn{3}{c}{\textbf{PyBullet (gpt4o)}} 
    & \multicolumn{3}{c}{\textbf{PyBullet (Gemini)}}\\
    \cmidrule(lr){2-4} \cmidrule(lr){5-7} \cmidrule(lr){8-10} \cmidrule(lr){11-13} \cmidrule(lr){14-16}
    \textbf{Agent} 
    & \textbf{SR$\uparrow$} & \textbf{GCR$\uparrow$} & \textbf{SRep$\uparrow$} 
    & \textbf{SR$\uparrow$} & \textbf{GCR$\uparrow$} & \textbf{SRep$\uparrow$} 
    & \textbf{SR$\uparrow$} & \textbf{GCR$\uparrow$} & \textbf{SRep$\uparrow$}
    & \textbf{SR$\uparrow$} & \textbf{GCR$\uparrow$} & \textbf{SRep$\uparrow$}
    & \textbf{SR$\uparrow$} & \textbf{GCR$\uparrow$} & \textbf{SRep$\uparrow$} \\
    \midrule
    Oracle
    & 33.04 & 50.04 & -
    & 31.65 & 51.71 & -
    & 50.00 & 85.75 & -
    & 19.00 & 32.50 & - 
    & 19.00 & 32.50 & - \\
    O-Ra 
    & 34.38 & 51.19 & 7.69
    & 34.17 & 54.08 & 14.16
    & 50.00 & 84.33 & 0.00
    & 53.00 & 61.58 & 39.13
    & 56.00 & 73.67 & 46.39 \\
    O-ERa 
    & 40.18 & 56.40 & 37.08
    & 42.81 & 61.81 & 33.33
    & 50.00 & 85.92 & 0.00
    & 75.00 & 80.50 & 71.24 
    & 72.00 & 78.25 & 65.65 \\
    O-LRa 
    & 34.38 & 51.00 & 6.73
    & 33.45 & 53.57 & 11.01
    & 93.00 & 97.04 & 87.00
    & \textbf{79.00} & \textbf{83.33} & \textbf{73.71}
    & \textbf{87.00} & \textbf{92.92} & \textbf{85.81} \\
    Baseline
    & 33.04 & 50.15 & 3.15
    & 32.01 & 51.89 & 0.98
    & 52.00 & 85.91 & 4.10
    & 67.00 & 74.92 & 59.06 
    & 82.00 & 89.08 & 78.90 \\
    O-LERa
    & \textbf{49.55} & \textbf{64.55} & \textbf{73.39}
    & \textbf{53.60} & \textbf{70.23} & \textbf{74.57} 
    & \textbf{94.06} & \textbf{98.17} & \textbf{95.03} 
    & 67.00 & 72.67 & 61.29 
    & 86.00 & 89.17 & 84.83 \\
    \bottomrule
\end{tabular}
}
\label{tab:lera_performance}
\end{table*}

\begin{table*}[th!]
\caption{
Performance of LERa with different VLMs across different environments.
}
\centering
\small
\resizebox{0.9\textwidth}{!}{
\begin{tabular}{lccc|ccc|ccc|ccc}
    \toprule
    & \multicolumn{3}{c}{\textbf{ALFRED-ChaOS (Seen)}} 
    & \multicolumn{3}{c}{\textbf{ALFRED-ChaOS (Unseen)}} 
    & \multicolumn{3}{c}{\textbf{VirtualHome-ChaOS}} 
    & \multicolumn{3}{c}{\textbf{PyBullet}} \\
    \cmidrule(lr){2-4} \cmidrule(lr){5-7} \cmidrule(lr){8-10} \cmidrule(lr){11-13}
    \textbf{VLM} 
    & \textbf{SR$\uparrow$} & \textbf{GCR$\uparrow$} & \textbf{SRep$\uparrow$} 
    & \textbf{SR$\uparrow$} & \textbf{GCR$\uparrow$} & \textbf{SRep$\uparrow$} 
    & \textbf{SR$\uparrow$} & \textbf{GCR$\uparrow$} & \textbf{SRep$\uparrow$} 
    & \textbf{SR$\uparrow$} & \textbf{GCR$\uparrow$} & \textbf{SRep$\uparrow$} \\
    \midrule
    LLaMA-3.2-11b
    & 35.71 & 52.01 & 11.11
    & 33.81 & 53.81 & 9.38
    & 52.00 & 74.17 & 20.00 
    & 34.00 & 45.92 & 20.54 \\
    LLaMA-3.2-90b
    & 38.84 & 54.80 & 30.58
    & 36.33 & 55.97 & 25.93
    & 54.00 & 84.25 & 8.00 
    & 64.00 & 71.08 & 57.68 \\
    Gemini-Flash-1.5
    & 46.43 & 61.61 & 67.22
    & 51.44 & 68.38 & 67.71
    & 59.40 & 72.51 & 24.00 
    & 55.00 & 66.33 & 46.75 \\
    Gemini-Pro-1.5
    & 42.19 & 56.51 & 56.16
    & 46.40 & 63.85 & 55.64
    & 65.35 & 87.87 & 41.58 
    & \textbf{86.00} & \textbf{89.17} & \textbf{84.83} \\
    gpt-4o-mini
    & 43.75 & 59.71 & 46.81
    & 46.04 & 63.97 & 49.12
    & 74.25 & 82.50 & 56.25 
    & 48.00 & 62.92 & 37.75 \\
    gpt-4o
    & \textbf{49.55} & \textbf{64.55} & \textbf{73.39}
    & \textbf{53.60} & \textbf{70.23} & \textbf{74.57}
    & \textbf{94.06} & \textbf{98.17} & \textbf{95.03} 
    & 67.00 & 72.67 & 61.29 \\
    \bottomrule
\end{tabular}
}
\label{tab:vlm_impact}
\end{table*}

\subsection{LERa Replanning Abilities}

Our experiments demonstrate that the LERa approach significantly improves task-solving success rates across all scenarios over Baseline. 
Table~\ref{tab:lera_performance} presents the results for Oracle, Baseline, LERa, and its variations.
In VirtualHome-ChaOS, LERa successfully solves almost all tasks, achieving 94\% SR. ALFRED-ChaOS, being a more complex environment than VirtualHome in terms of plan structure, has a more moderate improvement; however, the Oracle agent solves 50\% more tasks with LERa. In PyBullet, where a single action may fail even after replanning due to stochasticity, the impact of replanning is the most significant.

Experiments results with LERa variations~---~Ra, ERa, and LRa~---~demonstrate that most tasks can be solved without the ``Explain'' step in PyBullet scenarios. The main reason is the simplicity of PyBullet's actions. A detailed analysis is provided in Section~\ref{results:visual-feedback}. We were surprised to find that LRa solves tasks better than LERa in PyBullet. To verify the consistency of these results, we conducted additional experiments using the Gemini model.
The results from the PyBullet-Gemini experiments consistently indicate that replanning has a significant impact on task success rates.

\subsection{Visual Feedback Impact}
\label{results:visual-feedback}
Experiments in ALFRED confirmed that visual information is critical for understanding and correcting errors. However, tasks remain unsolvable without explicit failure reasoning and structured output formatting. The O-ERa agent fails to handle simple tasks due to an incomplete understanding of the environment, while O-LRa generates clear failure explanations but lacks stable and reasonable planning. Combining these approaches led to the LERa module and the O-LERa agent, which achieved the best results. The three-step replanning process, which includes extracting error-relevant information, reasoning about the failure, and performing format-constrained replanning, proved effective. Later experiments in VirtualHome confirmed these findings.

PyBullet is a more straightforward environment with fewer available actions and a shorter object list. The O-LRa achieved the best results, finally predicting stable and well-formatted action plans. Despite this, the three-step approach remains preferable due to its consistency and interpretability.

\begin{figure*}[ht!]
    \centering
    \includegraphics[width=\linewidth]{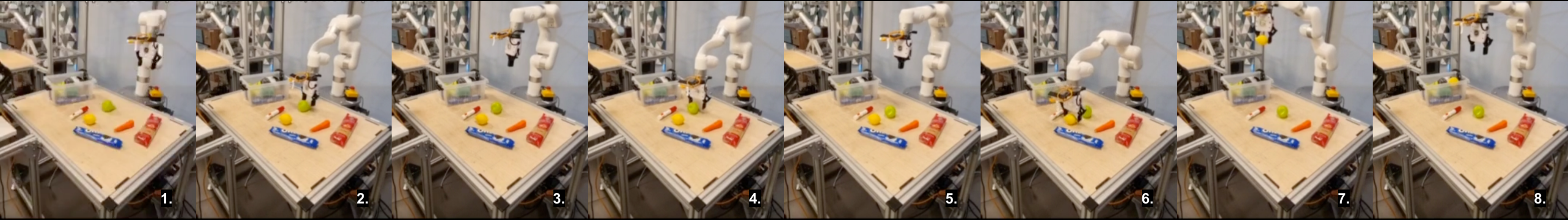}
    \caption{
    Hard replanning scenario in TableTop Robotic Stand experiments. Steps 2 and 4 mark error occurrences. In step 3, LERa intentionally repeats the previous action, attempting a successful retry. In step 5, after the pear moved slightly, the module successfully identified it as not an apple. Consequently, LERa excluded the ``pick apple'' action from the plan, as no apple was present on the table, and proceeded to execute the subsequent steps.
    }
    \label{fig:robot_run}
    \vspace{-10pt}
\end{figure*}

\subsection{Impact of Different VLMs}

The results of experiments with different VLMs are presented in Table~\ref{tab:vlm_impact}. 
Although LERa does not require training and operates using off-the-shelf VLMs, its performance is highly dependent on the accuracy of the VLM's image analysis.
A consistent decline in performance is observed across all environments as the quality of the VLM's picture analysis decreases.
Additionally, the choice of the VLM affects performance depending on the complexity of the environment. For simpler environments, such as PyBullet, Gemini models demonstrate superior results. In contrast, GPT-based models perform better in more complex environments that require advanced reasoning.

Although LERa involves multiple model queries, replanning is not required at every step, keeping the overall system load low. Latency can be further reduced through model inference optimization in future implementations.

\subsection{Impact of Non-Perfect Checker}
In real-world applications, designing a flawless action execution checker is challenging. In our series of additional experiments, we demonstrate how the LERa module performs when other modules make mistakes. Table~\ref{tab:not_ideal_checkers} shows that the use of the module significantly improves the success rate in all cases. Despite errors of the task checkers, LERa is able to cope with various situations not described in the few-shot prompts.

\begin{table}
\caption{The impact of the imperfect checker on replanning.}
\centering
\small
\begin{tabular}{lcccccc}
\toprule
& \multicolumn{3}{c}{\textbf{Seen split}} & \multicolumn{3}{c}{\textbf{Unseen split}} \\ \toprule
\textbf{Agent} & \textbf{SR$\uparrow$} & \textbf{GSR$\uparrow$} & \textbf{SRep$\uparrow$} & \textbf{SR$\uparrow$} & \textbf{GSR$\uparrow$} & \textbf{SRep$\uparrow$} \\
\midrule
O-15 & 12.50 & 35.90 & - & 12.95 & 39.75 & - \\
O-10 & 17.41 & 40.89 & - & 17.63 & 43.38 & - \\
O-05 & 24.55 & 45.01 & - & 24.82 & 47.54 & - \\
O-FC   & 33.04  & 50.04 & - & 31.65 & 51.71 & - \\ \midrule
O-L-15   & 22.32 & 44.42 & 35.48 & 20.50 & 45.29 & 25.00 \\
O-L-10   & 25.45 & 46.91 & 47.32 & 25.18 & 48.95 & 28.70 \\
O-L-05   & 34.38 & 52.57 & \textbf{54.16} & 35.97 & 55.76 & 35.43 \\
O-L-FC     & \textbf{44.64} & \textbf{59.00} & 52.83 & \textbf{46.76} & \textbf{63.46} & \textbf{49.34} \\ \bottomrule
\end{tabular}
\label{tab:not_ideal_checkers}
\vspace{-20pt}
\end{table}

\subsection{Results on the Real Robot}
To evaluate the performance of the LERa module, we conducted a series of experiments using the robotic arm tasked with manipulating various objects. The purpose of these experiments was to test replanning capabilities in scenarios with unexpected errors. A total of 18 series were conducted, each including different configurations of objects and targets. The agent successfully completed the replanning in 15 of the 18 series, demonstrating its reliability and adaptability under dynamic conditions.
Fig.~\ref{fig:robot_run} illustrates an example of a plan execution. The initial plan specified the task: ``Remove the apple and the lemon from the table''. The challenge in this scenario arose from a misclassification made by the detection module, which identified a green, round object~---~a pear~---~as an apple and incorporated it into the initial plan. Since the manipulator was unable to lift the pear, execution errors occurred. After the pear slightly rotated, the LERa module re-evaluated the object and correctly determined that it was not an apple, subsequently omitting the steps in the plan related to the misclassified object.

\subsection{Failure Analysis}
In ALFRED, one of the most common and significant challenges in replanning is the inability to infer the cause of failure from observations. Unlike in VirtualHome, where the agent’s camera is positioned slightly behind, providing a clear view of objects in most cases, ALFRED employs a first-person perspective. This often leads to ambiguity in perception, making it difficult for the replanner to determine whether the observed scene includes an open refrigerator, a microwave, or simply an empty shelf. The VirtualHome failure cases arose from similar problems, with the VLM misinterpreting the opened microwave door as a closed transparent one, likely due to its low position partially excluding the door from the frame. In PyBullet, most errors occurred when the VLM failed to predict the necessary ``locate object'' action before attempting to pick or place it, or when it did not fully resolve the initial instruction. The model could generate a new plan that corrected an observed drop error but failed to account for objects already placed in their correct positions. As a result, the agent incorrectly relocated blocks that were previously positioned correctly. This behavior was particularly noticeable in the outputs of smaller models, whereas GPT and Gemini demonstrated better understanding of why already placed blocks should not be moved.
Additionally, despite all random parameters being fixed, some intrinsic environmental stochasticity was observed. In certain cases, blocks that were randomly dropped fell into positions where the gripper was unable to pick them up, leading to further task failures.

\section{CONCLUSION}
\label{sec:conclusion}
Large Language Models excel in robotic task planning but struggle with dynamic changes and execution failures due to their reliance on textual input. To address this, we introduce LERa~---~a Visual Language Model-based replanning approach that leverages visual feedback for effective replanning. Unlike existing methods, LERa requires minimal additional information, making it a more practical and adaptable solution across different agent architectures.
We also propose three new environments, including ALFRED-ChaOS and VirtualHome-ChaOS, to evaluate replanning in dynamic settings. Extensive experiments show that LERa outperforms baselines in handling environmental and execution errors. Additionally, we integrate LERa into a tabletop manipulation robot, validating its effectiveness in real-world scenarios. Our findings highlight the robustness and generalizability of LERa. In future work, we will extend its capabilities to handle a broader range of errors, especially in long-horizon tasks requiring complex temporal and spatial reasoning. We believe that our work not only advances VLM-based replanning but also provides valuable benchmarks for future research in robotic task execution.

\section*{ACKNOWLEDGMENTS}
The study was supported by the Ministry of Economic Development of the Russian Federation (agreement with MIPT No. 139-15-2025-013, dated June 20, 2025, IGK 000000C313925P4B0002).

\bibliographystyle{IEEEtran}
\bibliography{IEEEexample}
\end{document}